\documentclass[runningheads]{llncs}

\usepackage{eccv}

\usepackage{eccvabbrv}

\usepackage{graphicx}
\usepackage{booktabs}
\usepackage{algorithm}
\usepackage{algpseudocode}
\usepackage{amsmath}
\usepackage{amssymb}
\usepackage{multirow}
\usepackage{makecell}
\usepackage{xcolor}
\usepackage[table]{xcolor}
\usepackage{multirow}
\usepackage{adjustbox}  
\usepackage{wrapfig, lipsum, booktabs}
\usepackage[accsupp]{axessibility}  
\definecolor{lightblue}{RGB}{230, 240, 250}

\usepackage[pagebackref,breaklinks,colorlinks,citecolor=eccvblue]{hyperref}

\usepackage{orcidlink}

\begin{document}

\title{PortraitGen: Exemplar-Driven GRPO with Dual-Reward Guidance for Photorealistic Portrait Generation}

\titlerunning{PortraitGen}

\author{Xiaomin Li~\textsuperscript{1~$\star$} \and
Qian Liang~\textsuperscript{2} \and
Yinan Li~\textsuperscript{3} \and
Ying Zhang~\textsuperscript{4} \and 
Chen Li~\textsuperscript{4} \and \\
Jing Lyu~\textsuperscript{4} \and
Huchuan Lu~\textsuperscript{1} \and
Xu Jia~\textsuperscript{1}}

\authorrunning{X.~Li et al.}

\institute{Dalian University of Technology \and
University of Electronic Science and Technology of China \and
Zhejiang University \and
WeChat, Tencent Inc. \\
\email{xmli22@mail.dlut.edu.cn},
\email{\{yinggzhang, chaselli\}@tencent.com},
\email{xjia@dlut.edu.cn}}

\maketitle
\let\thefootnote\relax\footnotetext{$^\star$ Work done during an internship at Tencent.} 

\begin{abstract}
\label{sec:abs}
Reinforcement Learning like Group Relative Policy Optimization~(GRPO) has significantly advanced text-to-image post-training. However, current methods often favor superficial aesthetics, such as over-saturated colors, leaving critical flaws like AI artifacts and biological implausibilities unresolved.  
We attribute these limitations to two primary factors: (1) The absence of real images during post-training confines GRPO sampling to the original distribution, failing to break inherent generative boundaries; (2) the optimization process lacks specific rewards targeting fine-grained artifacts like overly oily skin and other AI artifacts.
To address this, we propose PortraitGen, a novel framework tailored for photorealistic portrait generation. First, we break inherent generative boundaries by directly introducing real images into the GRPO sampling groups, where image inversion is employed to obtain their transition probabilities and latents. Second, to explicitly steer the model toward photorealism, we introduce a complementary dual-reward mechanism: OmniReward for general quality and AI-Portrait for human-centric fidelity. Furthermore, we curate PortraitBench, a comprehensive portrait-centric benchmark. Extensive experiments demonstrate that PortraitGen significantly outperforms existing baselines, effectively suppressing AI artifacts and achieving unprecedented photorealism.

\end{abstract}
\section{Introduction}
\label{sec:intro}
Recent advancements in text-to-image~(T2I) generation have underscored the pivotal role of reinforcement learning~\cite{ddpo,dpok,diffusiondpo,draft,flow-grpo,dancegrpo}. 
Built upon pre-trained models, these frameworks can further calibrate generation outputs toward human preferences defined by reward models~\cite{Pick-a-pic,imagereward,hpsv2,hpsv3}. 
Specifically, several methods based on group relative policy optimization~(GRPO)~\cite{deepseekmath}, such as FlowGRPO~\cite{flow-grpo} and DanceGRPO~\cite{dancegrpo}, have effectively optimized flow matching-based models (e.g., FLUX~\cite{flux-2,flux2024}) by integrating HPSv2.1~\cite{hpsv2} and CLIP~\cite{clip} scores as reward signals.

However, while these post-training techniques yield visually appealing images, they often induce superficial stylistic shifts, such as over-saturated colors, instead of rectifying underlying structural flaws. In the specific domain of photorealistic portrait generation, these methods struggle to mitigate the AI artifacts and address the lack of physical fidelity. Existing portrait generation works~\cite{UniPortrait,photomaker} predominantly focus on identity customization, often neglecting limb generation and biological implausibilities such as human with six fingers. Despite the widespread use of Reinforcement Learning~(RL) in general T2I tasks~\cite{t2ir1,mmr1}, there are few works tailored for human subjects that simultaneously optimize facial aesthetics and extremity integrity.

Recent efforts have begun to pivot toward portrait realism. \cite{realgen} identifies that models like FLUX~\cite{flux2024} and Qwen-Image~\cite{qwenimage} often tend to produce over-smoothed, unnatural oily skin. However, by relying on deepfake detection tools, \cite{realgen} merely classifies images as ``real'' or ``AI-generated'' based on feature-level artifacts, lacking a specialized focus on human-centric nuances. Conversely, while \cite{humanaesexpert} evaluates higher-level aesthetic dimensions~(e.g., facial beauty, physique, and environment), it remains aligned with traditional aesthetic standards and fails to quantify the AI artifacts or skin oiliness prevalent in flow matching generators. Consequently, there is an need for specialized reward models to explicitly quantify these AI artifacts. Beyond the lack of fine-grained reward signals, a more fundamental limitation of GRPO-based methods~\cite{flow-grpo,dancegrpo,pref-grpo} is the high similarity of sampled data within a group. Because samples are drawn from the same generative distribution, the model never observes authentic photographs during optimization, making genuine photorealism an unreachable goal. 

To this end, we propose PortraitGen, a post-training framework dedicated to suppressing AI artifacts and achieving superior photorealistic portrait generation. Specifically, to break inherent generative boundaries of traditional GRPO, we introduce real images directly into the sampling groups. 
To enable gradient backpropagation of the exemplar image, we perform image inversion to extract its requisite transition probabilities and latents across all sampling steps. 
However, relying solely on injected exemplars is insufficient, as the model lacks explicit feedback to guide the optimization. To actively propel the model toward genuine photorealism, we introduce two complementary reward models. On one hand, OmniReward provides point-wise evaluations across four fundamental dimensions: content, color/lighting, clarity, and composition. On the other hand, the AI-Portrait reward explicitly penalizes synthetic signatures, i.e., the AI artifacts and skin oiliness, employing pairwise win-rate scoring within groups to amplify subtle perceptual differences.
Additionally, we introduce PortraitBench, a comprehensive benchmark specifically tailored for portrait-centric assessment, thereby bridging the existing gap in evaluating portrait generation.
Extensive experiments demonstrate that our proposed Exemplar-Driven GRPO training framework significantly enhances the photorealism of portrait generation. 
The primary contributions of this work are as follows:
\begin{itemize}
    \item We introduce a novel exemplar-driven GRPO training paradigm that incorporates real images into the sampling group, effectively breaking the inherent generative boundaries of standard GRPO methods.
    \item We design two complementary reward models, OmniReward for general generation and AI-Portrait reward for human-centric fidelity, serving as both optimization signals and independent evaluation metrics.
    \item We curate PortraitBench, a diverse benchmark spanning various demographics, to comprehensively evaluate portrait realism across rich scenarios.
\end{itemize}

\section{Related Work}
\subsection{Text-to-image Generation}
Recently, T2I generation models~\cite{gpt5,seedream4,z_image} have witnessed remarkable advancements, serving as foundational tools for a diverse range of downstream tasks~\cite{Renaissance_survey,diffusion_survey,diffusion_survey2,t2i_survey}. 
Among these, generalized diffusion models~\cite{flux2024,flux-2,seedream4,sdxl} demonstrate superior efficacy in modeling the intrinsic distributions of complex, high-dimensional data manifolds.
Whether adopting stochastic diffusion processes or flow matching paradigms, these models establish a mapping from a tractable prior distribution $p(z) = \mathcal{N}(0, I)$ to the target data manifold by learning to reverse a forward noise injection trajectory~\cite{ddpm,ddim,Dpm-solver}. 
Beyond these pure generative models, unified multimodal architectures leverage autoregressive objectives to achieve seamless content understanding and generation; for instance, models such as Chameleon~\cite{chameleon}, Show-o~\cite{show_o}, OmniGen~\cite{Omnigen}, and Bagel~\cite{bagel} have exhibited potent text-to-image capabilities. In this work, we mainly focus on the post-training optimization of generalized diffusion-based models.

\subsection{Reward Models for Image Generation}
Reward models are essential for ensuring that generative models align with human intent~\cite{unifiedreward,reward_survey1,reward_survey2}. Although traditional metrics like FID~\cite{fid} and CLIP score~\cite{clip} provide a baseline for evaluating visual quality and semantic consistency, they often fail to capture complex human preferences. To mitigate this, recent approaches such as PickScore~\cite{Pick-a-pic} and HPSv2~\cite{hpsv2}) utilize human preference data to fine-tune CLIP~\cite{clip} encoders or train dedicated classifiers~\cite{laion5b}. Moreover, the emergence of Multimodal Large Language Models~(MLLMs)~\cite{qwen3vl,llama3,blip2} has introduced powerful backbones for reward modeling. These architectures~\cite{unifiedreward,hpsv3} are typically adapted using regression heads optimized with Bradley-Terry loss~\cite{imagereward}, or employed for direct pairwise image comparisons~\cite{viescore,LLaVA-Critic,unifiedreward}.
Despite these advancements, current models lack the sensitivity to assess portrait images or detect the specific synthetic artifacts associated with AI generation. To bridge this gap, we formulate two complementary reward models specifically tailored to evaluate human-centric fidelity and penalize AI artifacts. 

\subsection{Fine-tuning T2I Models with Rewards}
Recent T2I advancements focus on objective alignment via Reinforcement Learning~(RL) \cite{t2ir1,mmr1,rl_survey1}.
Several approaches optimize base models through direct reward maximization \cite{reneg,imagereward,textcraftor}.
More recently, Flow-GRPO~\cite{flow-grpo} and DanceGRPO~\cite{dancegrpo} instantiate GRPO within flow matching frameworks by reformulating deterministic Ordinary Differential Equations (ODEs) into equivalent Stochastic Differential Equations (SDEs), thereby introducing stochasticity to enable effective policy learning.
Alternatively, Pref-GRPO~\cite{pref-grpo} utilizes pairwise preference fitting to mitigate reward hacking. However, this method is not specifically tailored for portrait generation and falls short in resolving the prevalent AI artifacts. 
RealGen~\cite{realgen} employs synthetic image detection models to steer the generation process away from artificial artifacts at both semantic and feature levels. Although it alleviates skin oiliness, it struggles with structural distortions.
These models have achieved notable performance improvements. However, existing GRPO sampling schemes remain fundamentally limited, as the model never observes photographic-quality images during the sampling process. To effectively alleviate the persistent AI artifacts in current generative models and resolve the limitations of existing GRPO strategies, we propose a novel GRPO strategy.

\begin{figure}[t]
  \centering
  \includegraphics[width=1\linewidth]{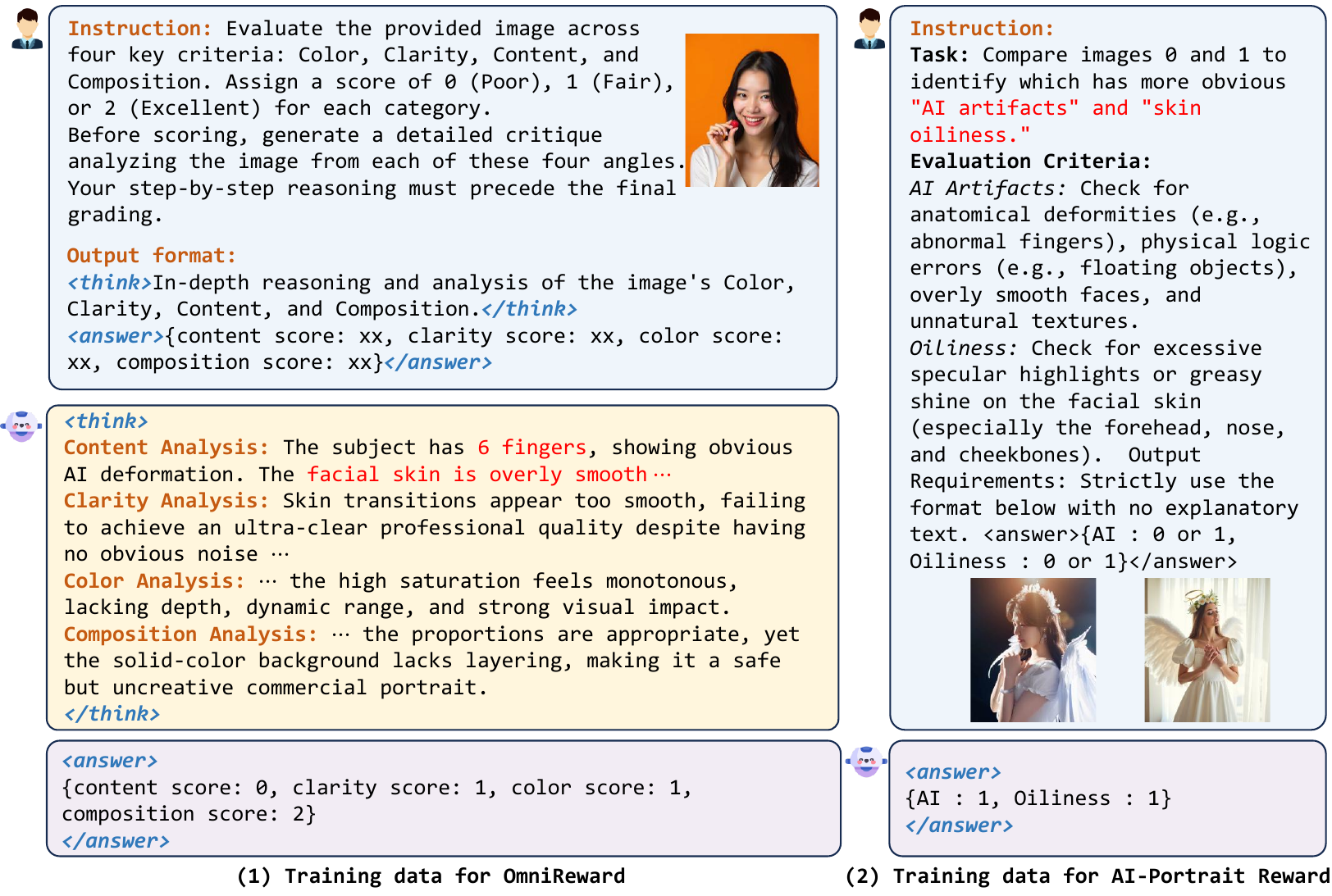}
  \caption{Training data samples for OmniReward and AI-Portrait Reward. ``Instruction'' denotes the user prompt, with <think> and <answer> tags enclosing the assistant's response. The reasoning process in OmniReward is truncated here for brevity.}
  \label{fig:reward_visulization}
\end{figure}
\section{Portrait-Centric Reward Models}
In this section, we present two complementary reward models to steer the model toward genuine photorealism during post-training. Specifically, OmniReward serves as a versatile point-wise evaluator, quantifying visual quality across four fundamental dimensions to reflect its ``Omni'' nature. Complementing this, we introduce AI-Portrait reward, a specialized pairwise ranker, to explicitly penalize AI artifacts prevalent in current generative models.

\subsection{OmniReward}
\noindent
\textbf{Data Annotation and Filtering.}
We curate a collection of high-quality, real-world portraits sourced from social platforms, primarily comprising celebrity street photography and editorial magazine captures, to ensure a high aesthetic baseline. Professional annotators are commissioned to evaluate each image across four distinct dimensions: Content, Clarity, Lighting and Color, and Composition. Each dimension is rated on a scale from 0 to 2, corresponding to ``Poor,'' ``Fair,'' and ``Excellent,'' respectively.
Beyond assigning numerical scores, annotators are required to provide explicit justifications for any zero-point ratings. For example, within the Content dimension, an image automatically receives a 0 if it presents mosaics, offensive content, or AI artifacts. Additional scoring criteria are detailed in the appendix.
Beyond establishing these strict evaluation rules, we also ensure the structural balance of the dataset.
To mitigate bias during model training, we balance the training data across the three discrete scores for each dimension. Recognizing the critical role of Content and Clarity, we explicitly prioritize a uniform distribution for these two dimensions to prevent class imbalance. The resulting processed training set yields over 90,000 high-quality samples.

\noindent
\textbf{Construction of Thinking Process.}
The fundamental objective of this reward model is to regress a quantitative score for a given image. To endow this scoring process with explicit interpretability, our constructed training data incorporates the underlying reasoning processes alongside the standard image-score pairs.
Specifically, human evaluators have previously annotated each image with explicit descriptive tags. By utilizing these human-provided tags as semantic anchors and grounding the reasoning process in direct visual evidence, an MLLM can successfully construct more exhaustive narratives that systematically articulate both synthetic flaws and inherent visual strengths. During training, these rationales serve as an explicit reasoning trace. An example of the training data is visualized in the first panel of Fig.~\ref{fig:reward_visulization}~(1).

\noindent
\textbf{Reward Model Training.}
We utilize Qwen3-VL-8B-Instruct~\cite{qwen3vl} as the backbone. Following the standard Supervised Fine-Tuning~(SFT) protocol, the model is optimized on our curated dataset via an autoregressive generative paradigm. Given an input image $x$ and the associated instruction $s$, the model learns to minimize the negative log-likelihood of the target response sequence $y$:
\begin{equation}
\mathcal{L} = - \mathbb{E}{(x, s, y) \sim \mathcal{D}} \left[ \sum_{t=1}^{T} \log P(y_t | s, x, {y}_{<t}; \theta) \right]
\end{equation}
\noindent
where $\theta$ denotes the learnable parameters and $y = \{y_1, y_2, \dots, y_{T}\}$ represents the serialized output sequence containing both the explicit reasoning trace and the final quantitative score.

\noindent
\textbf{Evaluation.} 
To verify the scoring precision of OmniReward, we construct a dedicated evaluation benchmark comprising 1,000 high-quality images paired with comprehensive ground-truth scores across the four dimensions.
As detailed in Tab.~\ref{tab:omnireward}, the fine-tuned OmniReward establishes a formidable performance margin over the vanilla Qwen3-VL-8B-Instruct model, achieving near-perfect accuracy across all dimensions. Remarkably, OmniReward even significantly surpasses the zero-shot performance of the much larger Qwen3-VL-235B-A22B-Instruct. These results underscore the efficacy of our fine-tuning in all complex dimensions, demonstrating that task-specific alignment fundamentally surpasses the raw capacity of scale-heavy foundational models.

\begin{figure}[t]
  \centering
  \includegraphics[width=0.8\linewidth]{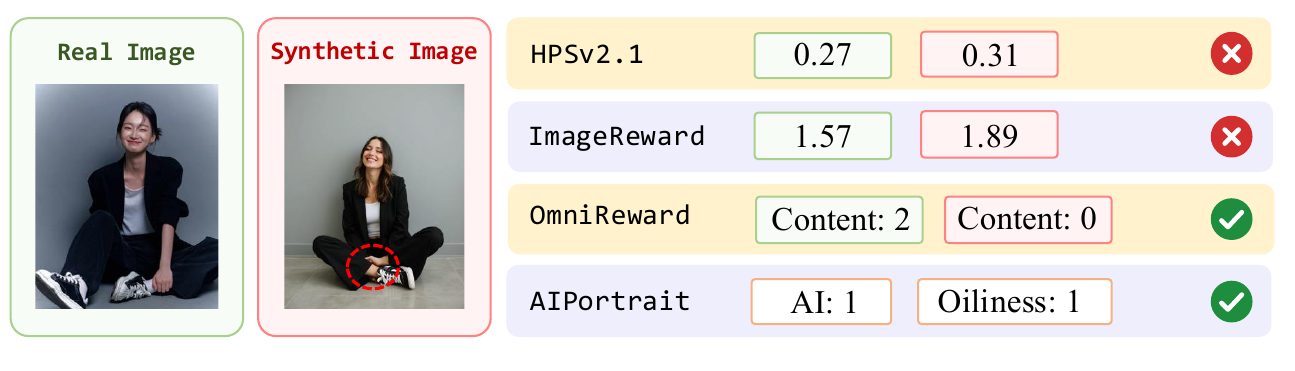}
  \caption{Quantitative scoring comparison between real and synthetic images using different reward models. Red dashed circles indicate severe structural distortions within the synthetic generation. Zoom in for best view.}
  \label{fig:reward_compare}
\end{figure}

\subsection{AI-Portrait Reward}

\noindent
\textbf{Data Construction.}
To establish a robust discriminative boundary between real and synthetic portraits, we construct a large-scale pairwise dataset. We initially curate approximately 10,000 real images. Conditioned on their original captions, we deploy FLUX.1-dev~\cite{flux2024}, Z-Image~\cite{z_image}, and Qwen-Image~\cite{qwenimage} to render corresponding synthetic counterparts. Pairing each real image with its three model-generated counterparts yields a total of 30,000 real-versus-synthetic image pairs. The formulated training data is structured into a standard user-assistant dialogue paradigm, tasking the model to select the sample manifesting more pronounced artificial characteristics, as illustrated in Fig.~\ref{fig:reward_visulization}~(2).
Unlike the strategy in OmniReward, we explicitly omit the deductive reasoning trajectory during this fine-tuning stage. Since images synthesized by generative models are not invariably of extremely poor quality, compelling the MLLM to explicitly annotate defects within them would likely induce model hallucinations, thereby introducing additional noise into the training process.
Consequently, rather than providing a thinking process, we solely assign the final score label. That is, we consider the synthesized image to possess a more pronounced AI artifacts and set its corresponding index as the ground-truth label.

\noindent
\textbf{Reward Model Training.}
The training pipeline for AI-Portrait Reward is analogous to that of OmniReward, where we fine-tune the Qwen3-VL-8B-Instruct. However, the target sequence $y = \{y_1, y_2, \dots, y_{T}\}$ is strictly confined to the final score label, omitting any intermediate reasoning. During training, the model is provided with image pairs and tasked with comparing the relative quality of AI artifacts and skin oiliness between the two images.
\begin{table}[t]
\centering
\caption{Prediction accuracy (\%) of OmniReward and AI-Portrait Reward compared with vanilla Qwen3-VL models on the test set. ``Avg.'' denotes the average accuracy.}
\label{tab:omnireward}
\resizebox{\linewidth}{!}{
\begin{tabular}{lcccccccc}
\toprule
\multirow{2}{*}{\textbf{Model}}
& \multicolumn{5}{c}{\textbf{OmniReward}}
& \multicolumn{3}{c}{\textbf{AI-Portrait Reward}} \\
\cmidrule(lr){2-6} \cmidrule(lr){7-9}
& \textbf{Content} & \textbf{Clarity} & \textbf{Color} & \textbf{Composition} & \textbf{Avg.}
& \textbf{AI-feel} & \textbf{Oiliness} & \textbf{Avg.} \\
\midrule
Qwen3-VL-8B-Instruct        & 13.50 & 1.30 & 8.20 & 3.90 & 6.73  & 51.25  & 50.80 & 51.03 \\
Qwen3-VL-235B-A22B-Instruct & 33.93 & 4.00 & 2.30 & 10.21 & 12.61 & 53.71  & 54.22    & 53.97 \\
\textbf{OmniReward}         & \textbf{86.20} & \textbf{95.20} & \textbf{95.80} & \textbf{85.70} & \textbf{90.73} & -   & -     & - \\
\textbf{AI-Portrait Reward} & -     & -     & -     & -     & -     & \textbf{96.2} & \textbf{96.5}   & \textbf{96.35} \\
\bottomrule
\end{tabular}
}
\end{table}

\noindent
\textbf{Evaluation.}
We construct an evaluation set consisting of 1,000 image pairs. Each pair comprises one real image and its corresponding generated counterpart. During inference, the presentation order within each pair is randomized to ensure an unbiased assessment. As summarized in Tab.~\ref{tab:omnireward}, our proposed method achieves over 96\% accuracy in identifying AI artifacts and skin oiliness. In contrast, the vanilla Qwen3-VL series baselines only achieve an accuracy of around 50\%, which is essentially equivalent to random guessing.

\noindent
\textbf{Comparison of Different Reward Models.}
As demonstrated in Fig.~\ref{fig:reward_compare}, we compare the evaluation scores from multiple reward models across real and synthesized image pairs. While the generated image contains obvious structural distortions in the hands, baseline reward models are unable to capture these localized failures, mistakenly predicting higher reward values for the synthetic sample. This critical oversight reveals that conventional human preference models are fundamentally ill-equipped to measure AI artifacts. Conversely, both OmniReward and the AI-Portrait reward possess the robust discriminative capacity to accurately assess content degradation and identify AI-generated signatures within the given image pair.

\section{Exemplar-Driven GRPO}
\subsection{Preliminary}
\noindent
\textbf{Group Relative Policy Optimization.}
The iterative image denoising process can be formulated as a Markov Decision Process~(MDP). Conditioned on a given prompt $c$, GRPO samples a group of $G$ rollouts $\{o_1, o_2, \dots, o_G\}$ from the old policy ${\pi}_{\theta_{old}}$. Subsequently, a reward model evaluates each rollout to assign a scalar reward $R_i$. Unlike standard reinforcement learning baselines that rely on an external value network for baseline estimation, GRPO leverages the group's internal statistics of the sampled group. Specifically, the advantage $\hat{A}^i_t$ for the $i$-th rollout at time step $t$ is conceptualized as the normalized score of the rewards strictly within this group, computed as follows:
\begin{equation}
\label{eq:advantage}
    \hat{A}^i_t = \frac{r_i - \text{mean}(\{r_i\}_{i=1}^G)}{\text{std}(\{r_i\}_{i=1}^G)}.
\end{equation}

\noindent
Finally, GRPO optimizes the policy model by maximizing the objective:
\begin{equation}
    \mathcal{J}_{GRPO}(\theta) = \mathbb{E}_{c \sim \mathcal{P}, \{o_i\}_{i=1}^G \sim \pi_{\theta_{old}}} \left[ f(r, \theta, \hat{A}) \right]
\end{equation}
where $\mathcal{P}$ represents the set of prompts, and the inner function is defined as:
\begin{equation}
    f(r, \theta, \hat{A}) = \frac{1}{G} \sum_{i=1}^G \frac{1}{|o_i|} \sum_{t=1}^{|o_i|} \left( \min \left( \rho^i_t \hat{A}^i_t, \text{clip}(\rho^i_t, 1-\epsilon, 1+\epsilon) \hat{A}^i_t \right) - \beta \mathbb{D}_{KL}(\pi_\theta || \pi_{ref}) \right)
\end{equation}

\noindent
Here, $\rho^i_t = \frac{\pi_\theta(o^i_{t-1}|o^i_t,c)}{\pi_{\theta_{old}}(o^i_{t-1}|o^i_t,c)}$ represents the importance sampling ratio between the current policy $\pi_\theta$ and the old policy $\pi_{\theta_{old}}$.

\noindent
\textbf{Image Inversion for Flow Matching.}
The primary objective of image inversion is to identify an initial noise map and its corresponding sampling trajectory, enabling the model to faithfully reconstruct the original image starting from this noise. This technique is widely adopted in image editing tasks. Traditional DDIM inversion~\cite{ddim} is primarily tailored for unconditional diffusion models. Furthermore, its forward and inversion processes rely on disparate iterative formulations, which often leads to significant discrepancies between the reconstructed sample and the original image.
To address this limitation, we adopt BELM (Bidirectional Explicit Linear Multistep method)~\cite{belm} to formulate a unified linear algorithm. By ensuring that both the forward and inversion trajectories strictly adhere to this identical formulation, our approach guarantees near-lossless image reconstruction. Specifically, given the sampling ODE $dx_t = v_\theta(x_t, t)dt$ in the flow matching framework, the forward and inversion processes are respectively formulated as follows:
\begin{align}
    \mathbf{x}_{i-1} &= \frac{h_i^2}{h_{i+1}^2} \mathbf{x}_{i+1} + \frac{h_{i+1}^2 - h_i^2}{h_{i+1}^2} \mathbf{x}_i - \frac{h_i(h_i + h_{i+1})}{h_{i+1}} v_\theta(\mathbf{x}_i, i) \\
    \mathbf{x}_{i+1} &= \frac{h_{i+1}^2}{h_i^2} \mathbf{x}_{i-1} + \frac{h_i^2 - h_{i+1}^2}{h_i^2} \mathbf{x}_i + \frac{h_{i+1}(h_i + h_{i+1})}{h_i} v_\theta(\mathbf{x}_i, i)
\end{align}
\noindent
where the step size $h_i$ is defined as $h_i = t_i - t_{i-1}$.

\subsection{Overall Post-training Framework}

\begin{figure}[t]
  \centering
  \includegraphics[width=1\linewidth]{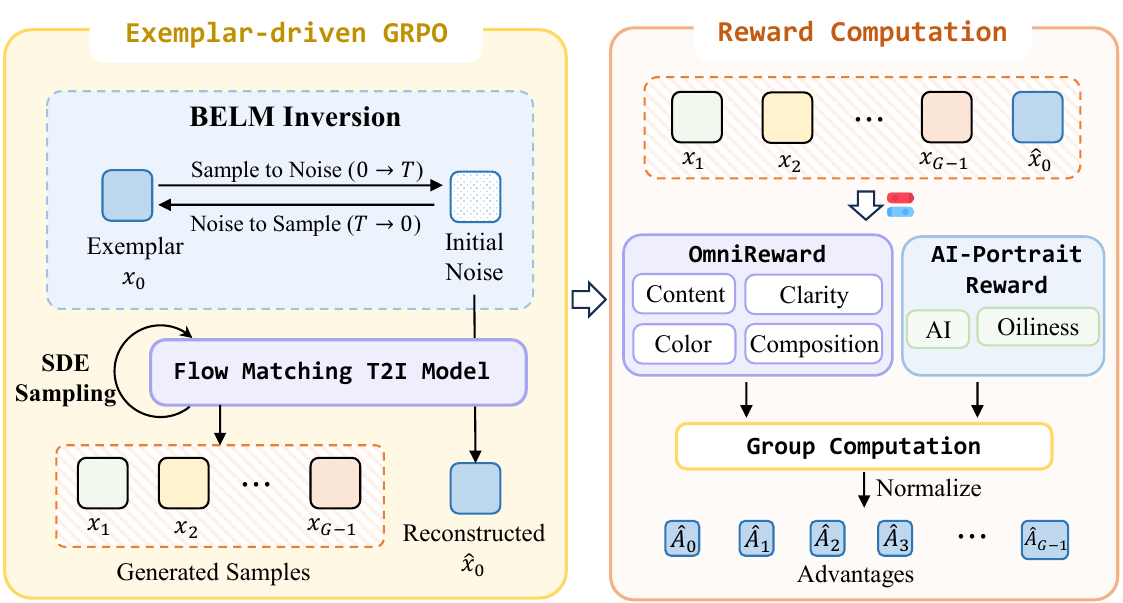}
  \caption{Overview of Exemplar-Driven GRPO. The T2I model generates $G-1$ images, which form a group alongside the exemplar image reconstructed via BELM inversion. Reward scores are then computed using OmniReward and AI-Portrait Reward. A gating mechanism is applied to selectively use these rewards or integrate new ones.}
  \label{fig:pipeline}
\end{figure}
To enable the sampling of diverse outputs within a single group under the GRPO framework, existing methods~\cite{flow-grpo,dancegrpo} inject stochasticity into the generation process. This is achieved by converting the deterministic probability flow ODE sampler into a reverse Stochastic Differential Equation~(SDE) that preserves an equivalent marginal distribution. However, relying entirely on the same generative model to sample all images within a group fundamentally fails to break through its inherent capacity ceiling. Because the T2I model has not genuinely observed real images during this phase, it inherently lacks the capability for true photorealistic generation.

Consequently, we introduce an advanced post-training paradigm as illustrated in Fig.~\ref{fig:pipeline}. In this formulation, we incorporate a real exemplar $\mathbf{x}_0$ into the optimization loop. Conditioned on the original prompt, multiple synthesized outputs alongside this real image collectively constitute a single group. However, incorporating this real image into the training pipeline is technically challenging. Although we can directly compute the reward for $\mathbf{x}_0$, model optimization remains impossible without the exact latent trajectories and corresponding step-wise prediction probabilities. Thus, it is necessary to reconstruct the image through resampling process that ensures strict consistency with the original real image. The most direct approach to achieve this is image inversion. Utilizing BELM, we accurately extract the intermediate latents and the step-wise trajectory probabilities of the exemplar image.

Ultimately, the overall training pipeline operates as follows. Given a group size $G$, the generative model synthesizes $G-1$ images through the reverse SDE, thereby injecting stochasticity during the sampling phase. The last image within the group is derived using the BELM algorithm via forward noise injection and subsequent reverse sampling, during which its trajectory prediction probabilities across discrete time steps are recorded. For all samples within this constructed group, we employ our OmniReward to evaluate each image across four dimensions: Content, Clarity, Lighting and Color, and Composition. Furthermore, the images within the group are exhaustively paired. Our AI-Portrait Reward evaluates each pair to determine which image exhibits a more pronounced synthetic appearance. The image displaying stronger generative artifacts is penalized as the loser, enabling the computation of a win-rate for each image. Formally, the win-rate $R_{win}(o_i)$ for the $i$-th image $o_i$ is formulated as:
\begin{equation}
R_{win}(o_i) = \frac{1}{G-1} \sum_{j=1, j \neq i}^G \mathbb{I}(o_i \succ o_j)
\end{equation}
where $\mathbb{I}(\cdot)$ denotes the indicator function, yielding 1 if $o_i$ contains fewer synthetic artifacts compared to $o_j$, and 0 otherwise.

\section{Experiments}

\subsection{Experimental Settings}
\noindent
\textbf{Training Dataset.}
To train our PortraitGen, we curate a training set of approximately 10,000 high-quality, photorealistic portrait images sourced from social media via rigorous manual screening. For annotation, Qwen3-VL-235B-A22B-Instruct is leveraged to generate keywords, short, and medium-to-long descriptions, which are randomly sampled at a ratio of 0.1:0.8:0.1 during training.

\begin{figure}[t]
  \centering
  \includegraphics[width=0.94\linewidth]{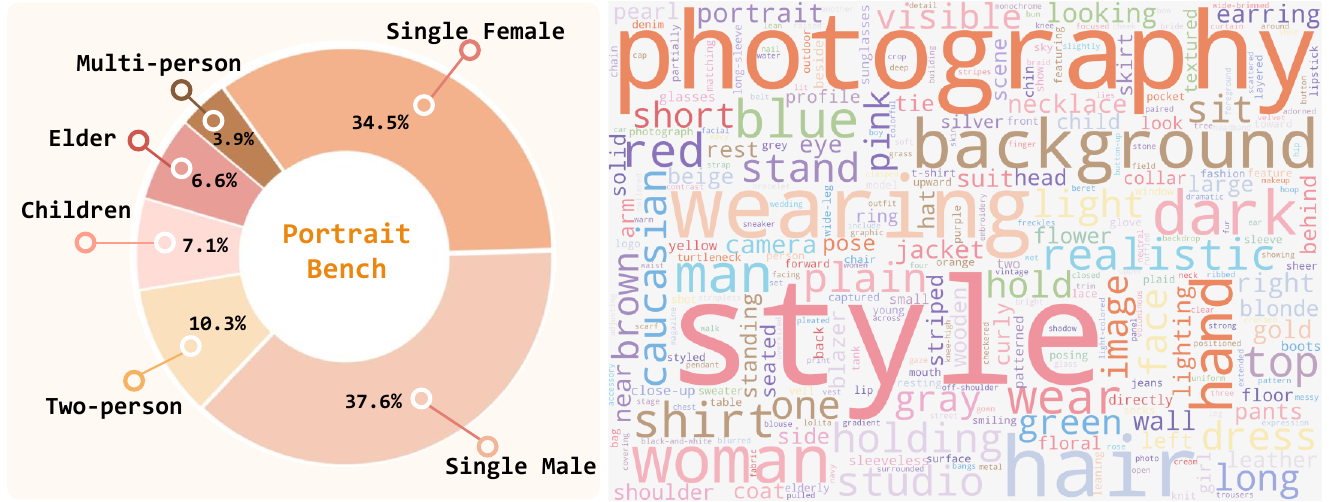}
  \caption{Distribution of our PortraitBench. Left: Categorical distribution of portrait scenarios. The benchmark encompasses diverse demographic groups, with percentages indicating their relative proportions. Right: Word cloud visualization of the benchmark.}
  \label{fig:benchmark}
\end{figure}

\noindent
\textbf{Evaluation Dataset.}
To systematically assess the model's capability in portrait generation, we construct a comprehensive evaluation benchmark. Specifically, we collect a wide variety of real human images and filter the data across key demographic attributes, including gender, age, and subject count. 
Ultimately, we obtain a benchmark comprising 1,000 structured samples, with its dataset distribution illustrated in Fig.~\ref{fig:benchmark}. As evidenced by the word frequency statistics, our benchmark places a deliberate emphasis on photorealistic portrait generation.

\noindent
\textbf{Implementation details.}
We adopt FLUX.1-dev as our baseline, strictly freezing the backbone to exclusively optimize the LoRA parameters. To satisfy the stringent numerical precision required by the BELM reconstruction, optimization is executed in full float32 with a learning rate of 5e-5, a mini-batch size of 1, and 12 gradient accumulation steps. During training, the sampling resolution structurally aligns with the group's exemplar image; however, to mitigate computational overhead, we constrain the maximum edge to 1024 pixels while preserving the original aspect ratio. The training phase employs a group size of 12 and 16 sampling steps, and is distributed across 16 NVIDIA H20 GPUs. For inference, we uniformly configure 28 sampling steps, a resolution of 1024, and a classifier-free guidance (CFG) scale of 3.5.

\noindent
\textbf{Metrics.} 
To comprehensively evaluate the generative capabilities of PortraitGen, we incorporate UnifiedReward~\cite{unifiedreward} and PickScore~\cite{Pick-a-pic} as additional evaluation metrics alongside our proposed OmniReward and AI-Portrait Reward models. This allows for a thorough comparative analysis with other leading methods. UnifiedReward employs MLLM to conduct a multi-dimensional assessment. Specifically, the Alignment metric evaluates semantic fidelity to the text prompt, Coherence penalizes structural distortions and visual artifacts to ensure logical consistency, and Style measures pure aesthetic quality independent of text relevance. Additionally, PickScore is deployed to quantify the overall generation quality and human preference conditioned on the respective text prompts.

\subsection{Qualitative Results}
\begin{figure}[!t]
  \centering
  \includegraphics[width=1\linewidth]{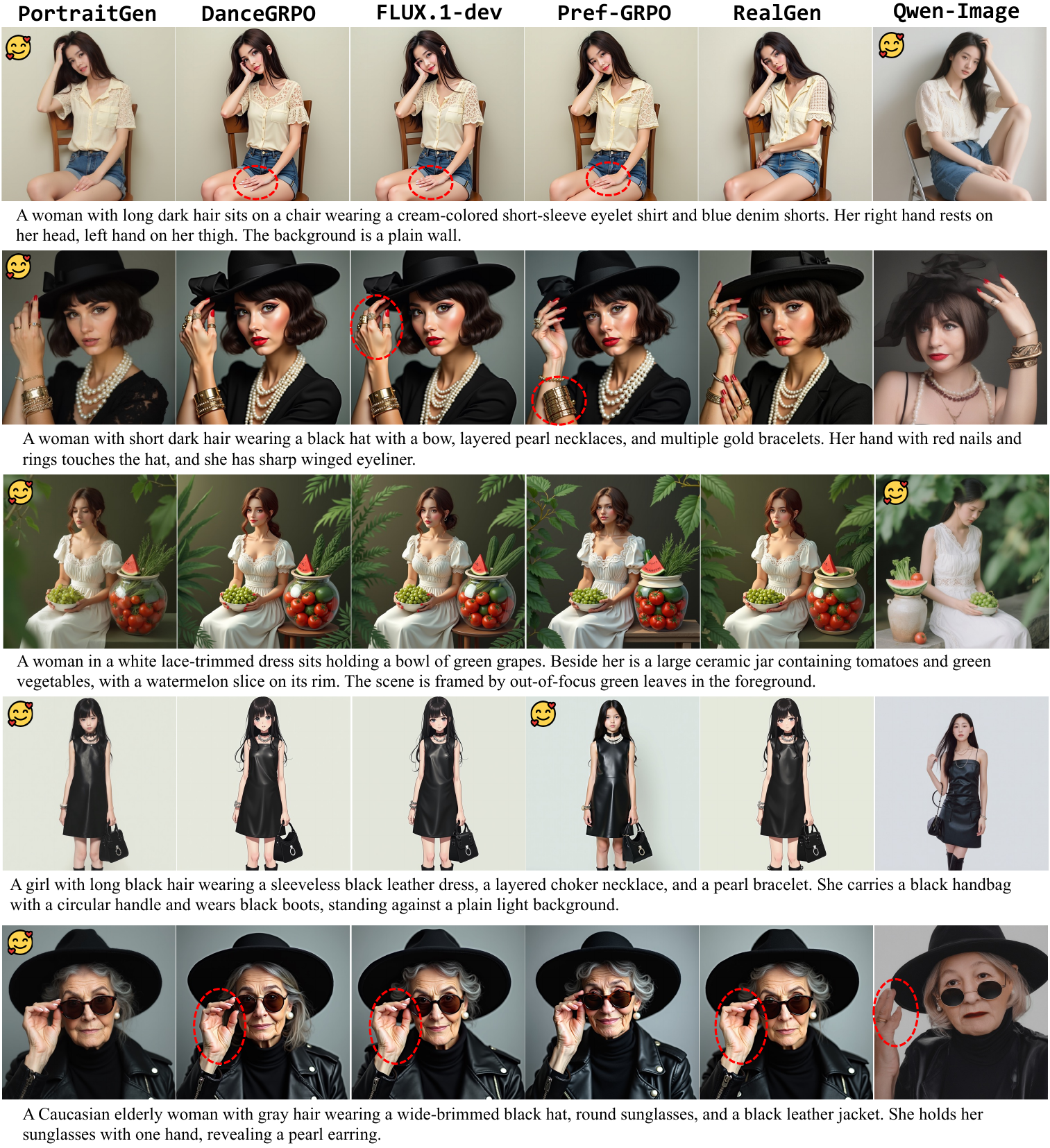}
  \caption{Qualitative comparisons between PortraitGen and other methods. The \includegraphics[height=1.0em]{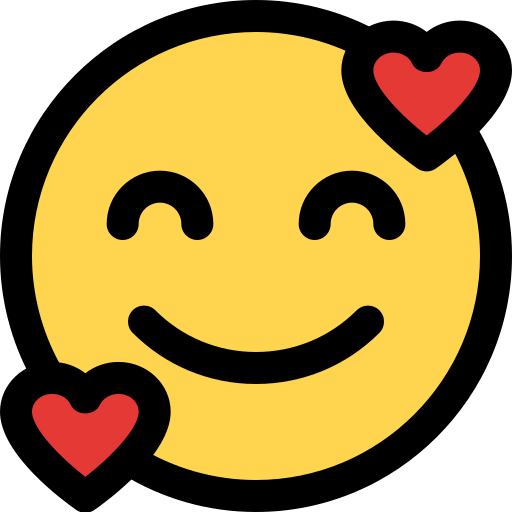} icon denotes the superior generated image for each specific text prompt. Red dashed circles highlight obvious structural distortions in the generated limbs.}
  \label{fig:qualititive}
  \vspace{-0.2cm} 
\end{figure}

To intuitively illustrate the comparative advantages of our method, we present a comprehensive qualitative evaluation. As shown in Fig.~\ref{fig:qualititive}, baseline models including FLUX.1 dev, DanceGRPO, and RealGen frequently default to oil painting or anime visual styles. Despite explicit instructions enforcing a photorealistic style within all prompts, these architectures fundamentally fail to generate realistic appearances. Conversely, Qwen-Image delivers exceptional aesthetic quality and avoids skin oiliness when generating Asian subjects. However, its outputs appear heavily processed by a distinctive stylistic filter. Furthermore, the model is noticeably less effective at generating subjects of diverse ethnicities, which predictably leads to visually chaotic content. Additionally, as evidenced by the fourth row of Fig.~\ref{fig:qualititive}, even when explicitly prompted to generate a young girl, Qwen-Image exhibits a persistent bias towards synthesizing adult females. Apart from our PortraitGen and Qwen-Image, all other evaluated methods exhibit severe skin oiliness. Ultimately, our proposed approach substantially suppresses common AI artifacts, particularly oily skin and structural limb distortions. Further comparative results are detailed in the appendix.

\subsection{Quantitative Results}
\begin{figure}[t]
    \centering
    \begin{minipage}{0.48\linewidth}
        \centering
        \includegraphics[width=\linewidth]{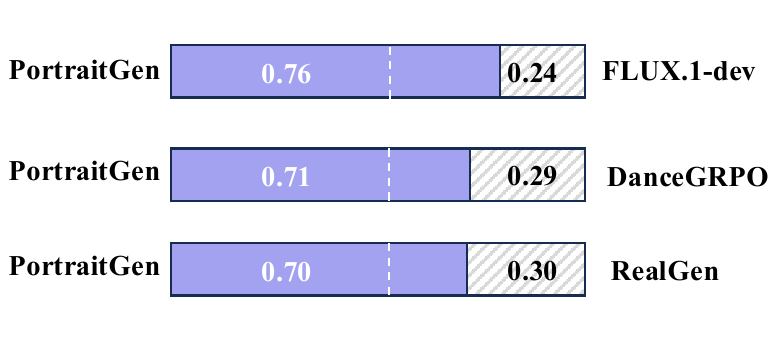}
        \caption{Win rate comparisons of PortraitGen against baseline models, evaluated using the AI Portrait metric.}
        \label{fig:ai_reward}
    \end{minipage}
    \vspace{-0.3cm} 
    \hfill
    \begin{minipage}{0.48\linewidth}
        \centering
        \includegraphics[width=\linewidth]{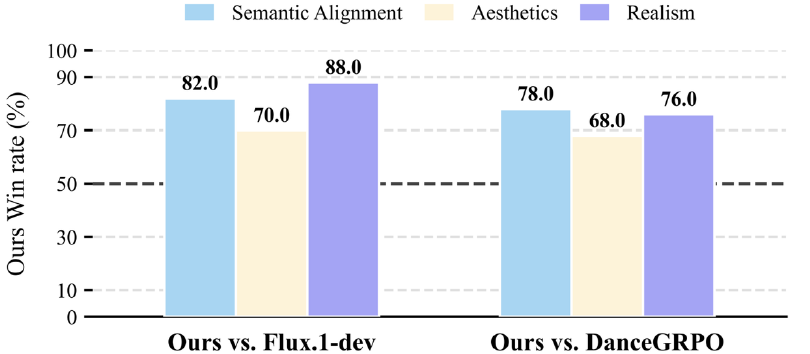}
        \caption{User study. Win rate comparisons between our method and baseline models across three evaluation dimensions.}
        \label{fig:user_study}
    \end{minipage}
    \vspace{-0.3cm} 
\end{figure}

The comparative baselines are primarily divided into base T2I models and post-training methods. To ensure a fair comparison, all post-training baselines employ FLUX.1-dev LoRA fine-tuning and undergo complete retraining on our benchmark. As shown in Tab.~\ref{tab:zhubiao}, PortraitGen consistently delivers the highest or second highest scores across the majority of criteria. Most notably, our method yields a significant enhancement in the OmniReward Content metric, a specific criterion for identifying synthetic traces. The highest Content value strongly validates that our strategy effectively mitigates intrinsic flaws within base T2I models, such as anatomical distortions and skin oiliness. Additionally, Fig.~\ref{fig:ai_reward} visualizes the win rate of our method against competing baselines under the AI Portrait reward. Even though baseline models occasionally synthesize good images that challenge direct pairwise discrimination, our framework consistently secures a superior overall win rate. This comprehensively confirms our profound capability to suppress AI artifacts.

\begin{table}[t]
\centering
\caption{Quantitative comparison of PortraitGen against leading baselines on our benchmark. The best and second-best results are highlighted in \textbf{bold }and \underline{underlined}, respectively. Higher values indicate better performance across all metrics.}

\label{tab:zhubiao} 
\resizebox{1\linewidth}{!}{ 
\begin{tabular}{lcccccccc}
\toprule
\multirow{2}{*}{\textbf{Method}}
& \multicolumn{4}{c}{\textbf{OmniReward}}
& \multicolumn{3}{c}{\textbf{UnifiedReward}} 
& \multirow{2}{*}{\textbf{PickScore}} \\
\cmidrule(lr){2-5} \cmidrule(lr){6-8} 
& \textbf{Content} & \textbf{Clarity} & \textbf{Color} & \textbf{Composition} & \textbf{Alignment} & \textbf{Coherence} & \textbf{Style}  \\
\midrule
\rowcolor{lightblue!50}
\multicolumn{9}{l}{\textit{Base T2I Models}} \\
\midrule
SDXL~\cite{sdxl} & 0.48 & \underline{0.99} & \textbf{0.97} & 1.20 & 3.08 & 3.68 & 3.26 & 22.16 \\
SD3.5-medium~\cite{sd3.5}     & \underline{0.94} & \textbf{1.00} & 0.94 & 1.19 & 3.43 & 3.66 & 3.19 & 22.17 \\
Bagel~\cite{bagel}            & 0.62 & 0.94 & \textbf{0.97} & 1.22 & \underline{3.50} & 3.71 & 3.25 & 22.66 \\
Qwen-Image~\cite{qwenimage}       & \textbf{0.97} & 0.95 & 0.94 & 1.19 & 3.38 & 3.62 & 3.14 & 21.69 \\
FLUX.1-dev~\cite{flux2024}       & 0.63 & 0.96 & \underline{0.95} & 1.25 & 3.48 & 3.79 & \underline{3.29} & 22.66 \\
\midrule
\rowcolor{lightblue!50}
\multicolumn{9}{l}{\textit{Post-training Methods}} \\
\midrule
DanceGRPO~\cite{dancegrpo}  & 0.65 & 0.95 & 0.93 & 1.25 & 3.48 & 3.80 & \textbf{3.30} & 22.64 \\
Pref-GRPO~\cite{pref-grpo}  & 0.61 & 0.97 & 0.94 & \underline{1.28} & \textbf{3.51} & 3.79 & 3.27 & \underline{22.71} \\
RealGen~\cite{realgen}      & 0.73 & 0.87 & 0.94 & 1.24 & 3.47 & \underline{3.81} & 3.26 & 22.53 \\
\rowcolor{lightblue!57}
\textbf{PortraitGen}          & \textbf{0.97} & 0.96 & \underline{0.95} & \textbf{1.30} & \textbf{3.51} & \textbf{3.83} & 3.27 & \textbf{22.77} \\
\bottomrule
\end{tabular}
} 
\end{table}

\subsection{Ablation Study}
To verify the effectiveness of our proposed method, we conduct ablation studies on PortraitGen to evaluate the individual contributions of the proposed reward models and the Exemplar-Driven GRPO. As detailed in Tab.~\ref{tab:xiaorongtable}, our complete approach achieves superior performance across most evaluation metrics. Notably, the Exemplar-Driven GRPO is designed to leverage real images as guidance, thereby encouraging the model to generate more diverse outputs, whereas the Dual-Reward module aims to accurately evaluate and score these varied results. Utilizing the reward models in isolation results in uniform scores for highly similar outputs, stalling meaningful knowledge acquisition and inevitably triggering reward hacking. Alternatively, relying exclusively on the Exemplar Driven GRPO promotes diversity, yet standard metrics like HPSv2.1~\cite{hpsv2} and CLIP~\cite{clip} lack the capability to accurately score these outputs, thereby bottlenecking further generative improvements. To resolve this, our comprehensive approach embeds the real exemplar directly into the sampled group. By leveraging our specialized reward models to measure the comparative advantage of this exemplar, we consistently steer the optimization trajectory toward the real image, culminating in photorealistic synthesis.

\begin{table}[t]
\centering
\caption{Ablation study. We evaluate the effects of integrating our proposed reward models and exemplar image into the DanceGRPO post-training architecture. Comp., Align., and Coher. denote Composition, Alignment, and Coherence, respectively.}
    \vspace{-4pt}
\label{tab:xiaorongtable}
\setlength{\tabcolsep}{3.5pt}
\resizebox{0.96\linewidth}{!}{ 
\begin{tabular}{lcc cccc ccc}
\toprule
\multirow{2}{*}{\textbf{Method}} 
& \multirow{2}{*}{\makecell{\textbf{Dual-}\\\textbf{Reward}}} 
& \multirow{2}{*}{\makecell{\textbf{Exemplar-}\\\textbf{Driven}}} 
& \multicolumn{4}{c}{\textbf{OmniReward}}
& \multicolumn{3}{c}{\textbf{UnifiedReward}} \\
\cmidrule(lr){4-7} \cmidrule(lr){8-10} 
& & & \textbf{Content} & \textbf{Clarity} & \textbf{Color} & \textbf{Comp.} & \textbf{Align.} & \textbf{Coher.} & \textbf{Style}  \\
\midrule
\rowcolor{lightblue!50}
\multicolumn{10}{l}{\textit{Base T2I Models}} \\
\midrule
FLUX.1-dev~\cite{flux2024} & - & - & 0.63 & \textbf{0.96} & \textbf{0.95} & 1.25 & 3.48 & 3.79 & 3.29 \\
DanceGRPO~\cite{dancegrpo} & - & - & 0.65 & 0.95 & 0.93 & 1.25 & 3.48 & 3.80 & \textbf{3.30} \\
\midrule
\rowcolor{lightblue!50}
\multicolumn{10}{l}{\textit{Ours Methods}} \\
\midrule
\multirow{3}{*}{\textbf{PortraitGen}}  & $\checkmark$ & $\times$ & 0.51 & \textbf{0.96} & \textbf{0.95} & 1.24 & 3.49 & 3.79 & 3.28 \\
 & $\times$ & $\checkmark$ & 0.61 & 0.92 & 0.94 & 1.24 & 3.49 & 3.78 & 3.29 \\
 & $\checkmark$ & $\checkmark$ & \textbf{0.97} & \textbf{0.96} & \textbf{0.95} & \textbf{1.30} & \textbf{3.51} & \textbf{3.83} & 3.27  \\
\bottomrule
\end{tabular}
}
\end{table}

\subsection{User Study}

We recruit five undergraduate students to participate in a user study. For this evaluation, we design a structured questionnaire consisting of 100 pairwise comparisons. Each trial presents participants with an image pair comprising a portrait generated by our PortraitGen alongside a corresponding output from either FLUX.1-dev or DanceGRPO. The evaluators are tasked with comparing the results and selecting the superior image across three critical dimensions: realism, aesthetics, and semantic alignment. 
As the results in Fig.~\ref{fig:user_study} demonstrate, our method consistently surpasses the baseline models across all three evaluation criteria. Most notably, regarding the realism dimension, our PortraitGen significantly mitigates prevalent AI artifacts.

\section{Conclusion}
\label{sec:conclusion}
In this paper, we propose PortraitGen to address AI artifacts, such as structural distortions and oily skin, in photorealistic portrait generation. 
By introducing an exemplar image into the traditional GRPO sampling process, we enable the model to perceive the authentic image distribution during training. Concurrently, we introduce the novel OmniReward and AI Portrait Reward models. These models accurately quantify the comparative advantage of the exemplar image over sampled group images. This mechanism compels the generative model to optimize toward the exemplar, effectively breaking the boundaries of conventional GRPO. Furthermore, we introduce PortraitBench, a comprehensive benchmark designed to evaluate generative models in photorealistic portrait synthesis. Despite current progress, reward models for measuring synthetic artifacts still possess room for improvement. In the future, we plan to collect broader datasets to train more robust reward models, extend our approach to temporally consistent video portrait generation, and incorporate fine grained spatial controls.



%
%
\bibliographystyle{splncs04}
\bibliography{main}
\end{document}